\documentclass[conference,a4paper]{APSIPA2025}
\usepackage{amsmath}
\usepackage{graphicx}
\usepackage{multirow}
\usepackage{threeparttable}
\usepackage[backend=biber,style=ieee,]{biblatex}
\usepackage{amsfonts}
\usepackage{booktabs}
\usepackage{subcaption}
\usepackage{geometry}
\usepackage{fancyhdr}

\fancypagestyle{firststyle}{
  \fancyhf{}
  \fancyhead[C]{2025 Asia Pacific Signal and Information Processing Association Annual Summit and Conference (APSIPA ASC)}
}

\begin{document}

\title{Mitigating Data Imbalance in Automated Speaking Assessment}

\author{
\authorblockN{
Fong-Chun Tsai \authorrefmark{1}, 
Kuan-Tang Huang \authorrefmark{1}, 
Bi-Cheng Yan \authorrefmark{1}, 
Tien-Hong Lo \authorrefmark{1}, and
Berlin Chen \authorrefmark{1}
}

\authorblockA{
\authorrefmark{1}
National Taiwan Normal University \\
E-mail: \{fongchun.tsai, 61347002s, 80847001s, teinhonglo, berlin\}@ntnu.edu.tw
}
}

\maketitle
\thispagestyle{firststyle}
\pagestyle{fancy}

\begin{abstract}
Automated Speaking Assessment (ASA) plays a crucial role in evaluating second-language (L2) learners’ proficiency. However, ASA models often suffer from class imbalance, leading to biased predictions. To address this, we introduce a novel objective for training ASA models, dubbed the Balancing Logit Variation (BLV) loss, which perturbs model predictions to improve feature representation for minority classes without modifying the dataset. Evaluations on the ICNALE benchmark dataset show that integrating the BLV loss into a celebrated text-based (BERT) model significantly enhances classification accuracy and fairness, making automated speech evaluation more robust for diverse learners.
\end{abstract}

\section{Introduction}
\label{sec:intro}

Driven by advances in automatic speech recognition (ASR) \cite{graves2006connectionist,NIPS2015_1068c6e4,pmlr-v202-radford23a} and the worldwide growth of second-language learners (L2), Automated Speaking Assessment (ASA) has emerged as a practical and widely used tool for language acquisition \cite{wang_towards_2018, banno_l2_2022, lo_effective_2024}, serving as a key component of computer-assisted language learning (CALL). ASA systems are developed to assess the spoken language proficiency of L2 learners in multiple dimensions of communicative competence, such as delivery, language use, and discourse coherence. Their applications primarily fall into two main areas: high-stakes exams and pedagogical tools. For example, organizations such as the ETS\cite{zechner_automated_2019} leverage ASA systems to streamline scoring processes, reduce the workload of human experts, lower recruitment costs, and deliver consistent and objective evaluations of speaking proficiency\cite{xu_assessing_2021}. On the other hand, ASA systems serve as a critical resource in classrooms, providing teachers with reliable and timely assessment feedback on student oral skills, in order to improve instructional quality and/or offer statistical data on student performance \cite{lo_effective_2024}.

A de facto standard for ASA systems is typically instantiated in an open-response learning scenario, where an L2 learner is provided with text prompts and/or questions to elicit natural speech responses. Based on open-ended and spontaneous speech inputs, ASA systems manage to quantify communicative competence from multiple facets, including pronunciation, vocabulary, grammatical accuracy, and topical relevance. Existing studies frame the ASA task as a proficiency score classification problem, which comprises scoring, speech processing, and natural language processing modules. Among these modules, the speech processing module is a critical component of ASA, acting as the foundation for analyzing the spoken responses of learners in terms of specific aspects, including pronunciation, stress, fluency, and intonation\cite{zechner_automated_2019}. There are two primary approaches to feature extraction: hand-crafted features and self-supervised learning (SSL) features \cite{hsu2021hubert, baevski2022data2vec}.  To extract hand-crafted features, the spoken response is first transcribed into a sequence of linguistic units by an ASR system, and a set of corresponding features is extracted to portray the learner’s pronunciation quality through forced alignment with the recognized hypotheses. The commonly used hand-crafted features include, but are not limited to, silence duration, speaking rate, and repetition rate. In contrast, recent advances in self-supervised models like Wav2Vec 2.0 \cite{baevski_wav2vec_2020} and BERT \cite{devlin_bert_2019} have introduced a new paradigm for feature extraction. These models leverage vast amounts of unlabeled data during pre-training to produce rich, high-dimensional representations of speech. While SSL models require fine-tuning to align with the specific goals of ASA and improve their interpretability, studies have shown that SSL features outperform traditional methods in ASA tasks \cite{banno_proficiency_2022}. This marks a significant step forward in enhancing the accuracy and robustness of speech assessments.

Most ASA data is derived from examination results, and the scores from these exams typically follow a normal distribution. This characteristic results in the ASA dataset often facing issues of imbalance, meaning that the number of data samples in each performance category or level within the dataset is uneven. Many studies in the past have pointed out that class imbalance can significantly affect the performance of classification tasks, leading to reduced accuracy and reliability \cite{buda2018systematic, johnson_survey_2019}. However, despite the well-recognized effects of data imbalance, there has not been much research or effort in previous ASA-related tasks to address or mitigate this issue effectively.

To address the issue of class imbalance, one common approach is data augmentation \cite{park19e_interspeech,shorten2019survey}. This can involve techniques such as reducing the number of instances in majority classes or generating additional data through interpolation methods. However, because ASA datasets have very limited amounts of data, reducing the number of majority class examples may lead to model overfitting.
Furthermore, since ASA data follows a Gaussian distribution, there is an inherent lack of examples at the high and low ends of the score range. This creates a problem when using interpolation-based data augmentation methods, as these methods rely on creating new synthetic examples by generating values between two data points. Therefore, this approach is also difficult to apply to ASA tasks.

To address the issue of class imbalance in ASA, we introduce Balancing Logit Variation (BLV) loss from the semantic segmentation domain \cite{wang_balancing_2023}, which does not require any modifications to the data. This method is particularly well-suited for ASA tasks that suffer from a lack of high- and low-score data and limited sample sizes. During training, this approach perturbs the model's predictions, causing each sample to be projected into a small range of feature space rather than being mapped to a single fixed feature point.

By projecting the data from minority classes into larger intervals, we effectively balance the feature space utilized by the model, thereby mitigating the data imbalance issue. This innovative strategy improves the model's ability to generalize across classes while addressing the challenges associated with imbalanced datasets in ASA tasks.

Our main contributions are threefold: 
\begin{enumerate}
\item \textbf{Introduction of BLV loss for ASA:} 
The paper proposes a novel loss function that perturbs the logits based on class frequency, enabling more effective learning from imbalanced datasets in ASA tasks.
\item \textbf{Integration into an ASA pipeline using Whisper and BERT:}
The method is integrated into a practical ASA pipeline using Whisper \cite{pmlr-v202-radford23a} for speech transcription and BERT for proficiency classification, showing its compatibility with state-of-the-art models.
\item \textbf{Empirical Validation with Extensive Experiments on Long-Tailed Speech Assessment Dataset:} 
Through extensive experiments, the paper demonstrates that BLV loss improves classification accuracy and fairness across both head and tail classes, outperforming baseline loss functions in real ASA scenarios.
\end{enumerate}

\section{Related Work}
\subsection{Feature Extraction}
ASA systems evaluate communicative competence by extracting features related to pronunciation, vocabulary, grammatical accuracy, and topical relevance. Traditional approaches relied on hand-crafted features to capture various aspects of language proficiency, including pronunciation, stress, fluency, and intonation \cite{coutinho_assessing_2016, Bhat_Yoon_2015}. However, these manual feature engineering methods often incorporate implicit assumptions that may inadvertently omit crucial proficiency indicators \cite{banno_l2_2022}.

Recent advancements in SSL have introduced more sophisticated methods for spoken response representation. Transformer models, trained through contrastive learning, have demonstrated remarkable capability in capturing multiple dimensions of speech, including semantics, acoustic properties, fluency, and pronunciation features \cite{singla_what_2022}. These models have achieved state-of-the-art performance across various downstream tasks. In the context of L2 learner assessment, several studies have successfully applied SSL features to ASA tasks \cite{banno_l2_2022, banno_proficiency_2022, lo_effective_2024}. Notably, Bannò et al. demonstrated that Wav2Vec2.0-based systems surpass BERT-based text-only approaches in proficiency assessment, particularly in evaluating speech-related components \cite{banno_proficiency_2022}. The superior performance of SSL features in ASA tasks can be attributed to their comprehensive representation of both semantic content and prosodic elements.

\subsection{Data Imbalance}

Data imbalance is a common classification problem that has been discussed in various fields. It leads to models that focus on the majority class, causing the performance of minority classes to degrade.

The previous literature has proposed several methods to address data imbalance, which can be mainly divided into two approaches. The first approach involves data augmentation techniques, such as oversampling, undersampling, or mixing. However, data augmentation can generate overly ambiguous samples, which, in tasks such as ASA with limited training data, could have a severe negative impact on model performance.

The second approach involves adjusting the loss function, such as using focal loss \cite{lin2017focal} or BLV loss. These methods are independent of the data and suitable for tasks such as ASA. In this work, we introduce the BLV loss from image segmentation. This loss introduces perturbations to the model predictions, allowing the features of minority classes to be projected onto a larger area, thus balancing the model performance.

\begin{figure}[ht]
    \centering
    \includegraphics[width=1\linewidth]{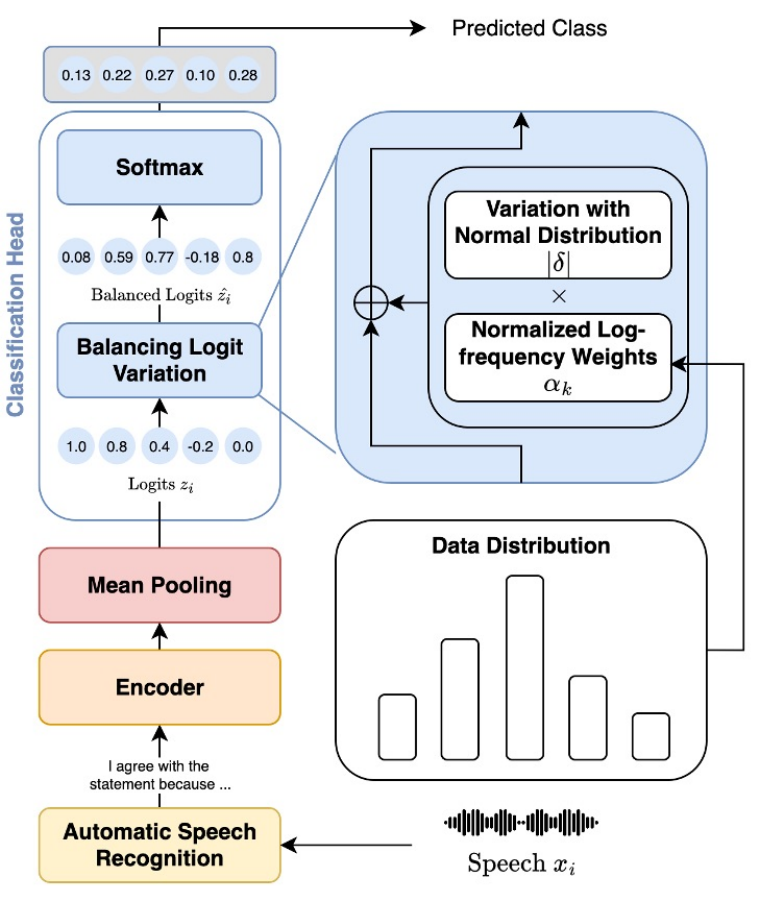}
    \caption{Model Architecture}
    \label{fig:model-architecture}
\end{figure}

\begin{table*}[!ht]
\centering
\setlength{\tabcolsep}{12pt}
\caption{Comparison of Models on ICNALE Dataset}
\begin{tabular}{l c c c c c c c c c}
\toprule
Method & $\sigma$ & PCC ↑ & \multicolumn{2}{c}{RMSE ↓} & \multicolumn{2}{c}{Accuracy (\%) ↑} & \multicolumn{2}{c}{Adjacent Accuracy (\%) ↑} & F1-score ↑ \\
\cmidrule(lr){4-5} \cmidrule(lr){6-7} \cmidrule(lr){8-9}
 & & & Standard & Macro & Standard & Macro & Standard & Macro & \\
\midrule
Bert (baseline) & --  & 0.6690 & 0.8614 & 0.9449 & 55.30 & 47.96 & 91.71 & 85.79 & 0.4672 \\
Bert + focal loss & -- & 0.6466 & 0.8877 & 0.9646 & 54.83 & 48.13 & 90.32 & 83.67 & 0.4828\\
\midrule
Bert + BLV & 2   & 0.6719 & 0.8587 & 0.9598 & \textbf{57.14} & 48.60 & 91.24 & 86.45 & \textbf{0.5000} \\
Bert + BLV & 6   & \textbf{0.6748} & \textbf{0.8479} & \textbf{0.9281} & 56.22 & \textbf{49.29} & \textbf{92.17} & \textbf{87.12} & 0.4952 \\
\bottomrule
\end{tabular}
\label{tab:ICNALE result}
\end{table*}

\section{Methodology}

\subsection{Model Overview}

Our model utilizes a pre-trained BERT encoder\cite{devlin_bert_2019} to generate representations for classification tasks. We use this model for simplicity to demonstrate the effectiveness of our proposed loss. Initially, the audio input is transcribed into text using the Whisper ASR system, which may introduce transcription errors. The resulting text is tokenized using a pre-trained BERT tokenizer.

Subsequently, the tokenized text is processed by the pre-trained BERT model, which serves as a feature extractor. 
The model obtains a fixed-length representation of the input sequence by applying mean pooling over the BERT embeddings. This representation is then fed into a feed-forward classification head that outputs logits for each class. The model architecture is illustrated in Fig. \ref{fig:model-architecture}.


In the classification phase, the contextualized representation is passed through a mean pooling layer to aggregate features, followed by a classifier head.

\subsection{BLV Loss}

When training on a long‐tailed dataset $\mathcal{D}=\{(\mathbf{x}_i,y_i)\}_{i=1}^N$ of spoken utterances $\mathbf{x}_i$ with labels $y_i\in\{1,\dots,C\}$, standard cross‐entropy tends to favor majority classes. To mitigate this, we adopt the \emph{Balancing Logit Variation} (BLV) strategy \cite{wang_balancing_2023}, which injects class‐dependent perturbations into the logits before computing the loss, where the perturbation for minority classes is larger and for majority classes is smaller.

Let $\mathbf{z}_i \in \mathbb{R}^C$ be the original C-dimensional logit vector for instance $i$, and let $q_k$ denote the total count of category $k$ with $N = \sum_{j=1}^C q_j$, Compute the normalized log-frequency weight

\begin{align}
\alpha_k = \dfrac{\log(N/q_k)}{\max_{j \in [C]} (\log(N/q_j))} \tag{1}
\end{align}

At each optimization step, sample $\boldsymbol{\delta}\sim\mathcal{N}(0,\sigma^2)$ and form the perturbed logit.

\begin{align}
    \mathbf{\hat z}_i^k = \mathbf{z}_i^k + \alpha_k\,|\boldsymbol{\delta}|\tag{2}
\end{align}

Here, $\sigma>0$ is the only BLV loss hyperparameter.
Next, compute class probabilities via softmax on $\{\mathbf{\hat z}_i\}$ and evaluate the cross-entropy loss.

\begin{align}
\mathcal{L}_{\mathrm{BLV}}
= -\frac{1}{N}\sum_{i=1}^N\sum_{k=1}^C \log\frac{\exp(\mathbf{\hat z}_i^k)}{\sum_{j=1}^C\exp(\mathbf{\hat z}_i^j)}\tag{3}
\end{align}

\section{Experimental Setups}

\begin{figure}[ht]
\centering
\includegraphics[width=1\linewidth]{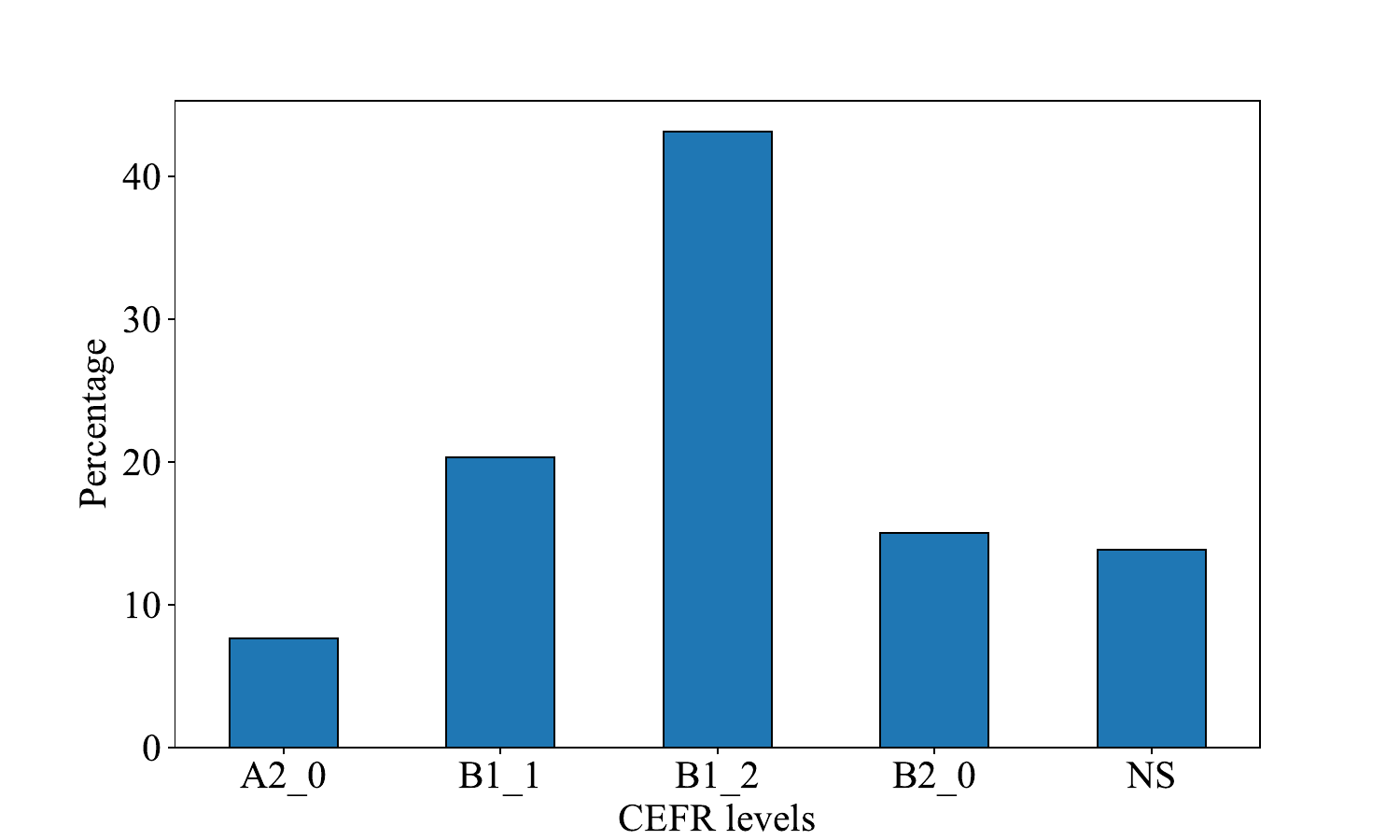}
\caption{Distribution of ICNALE Dataset on CEFR LEVEL}
\label{fig:ICNALE_distribution}
\end{figure}

\subsection{Dataset}
To validate the generalizability of our approach, we conducted experiments on the International Corpus Network of Asian Learners of English (ICNALE) corpus, a publicly available dataset \cite{lshikawa2023Icnale}.
We used the Spoken Monologues section of ICNALE, which contains a total of 4,332 samples. Each sample includes a topic, along with the user's spoken and written responses. The users consist of both native speakers and L2 learners whose CEFR (Common European Framework of Reference for Languages) levels\cite{division_common_2001} range from A2 to B2. The CEFR levels are determined based on the learners' vocabulary size and proficiency scores in English proficiency tests such as IELTS and TOEFL. As shown in Fig. \ref{fig:ICNALE_distribution}, there is an imbalance in the distribution of user levels within the dataset.


\begin{figure*}
    \centering

    \begin{subfigure}{0.4\textwidth}
        \centering
        \includegraphics[width=\textwidth]{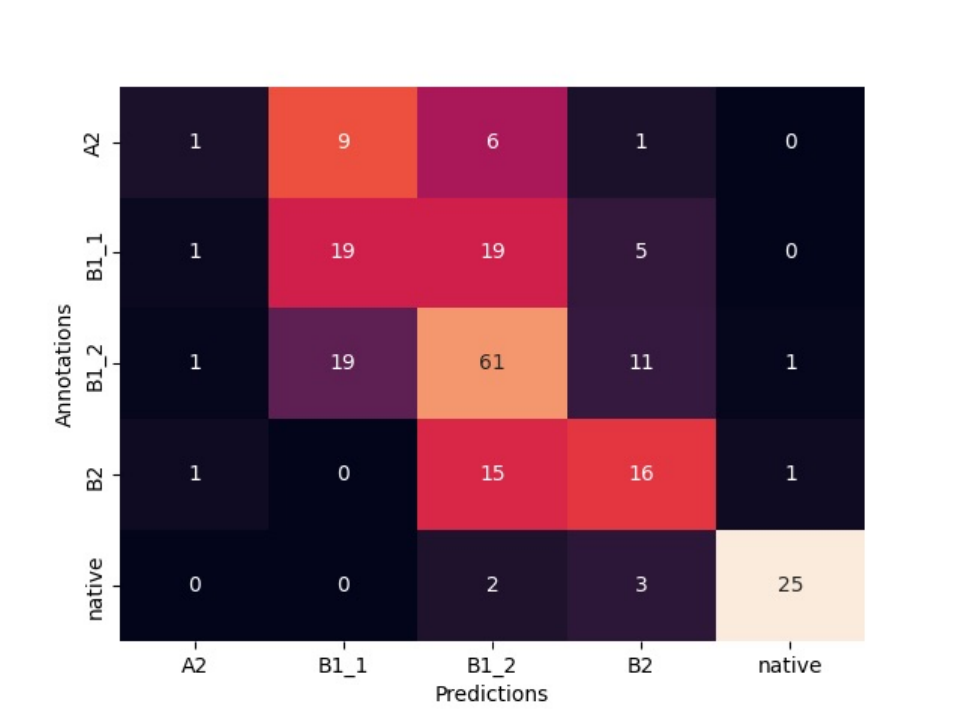}
        \caption{BLV $\sigma = 6$}
        \label{fig:blv-cm}
    \end{subfigure}
    \hfill
    \begin{subfigure}{0.4\textwidth}
        \centering
        \includegraphics[width=\textwidth]{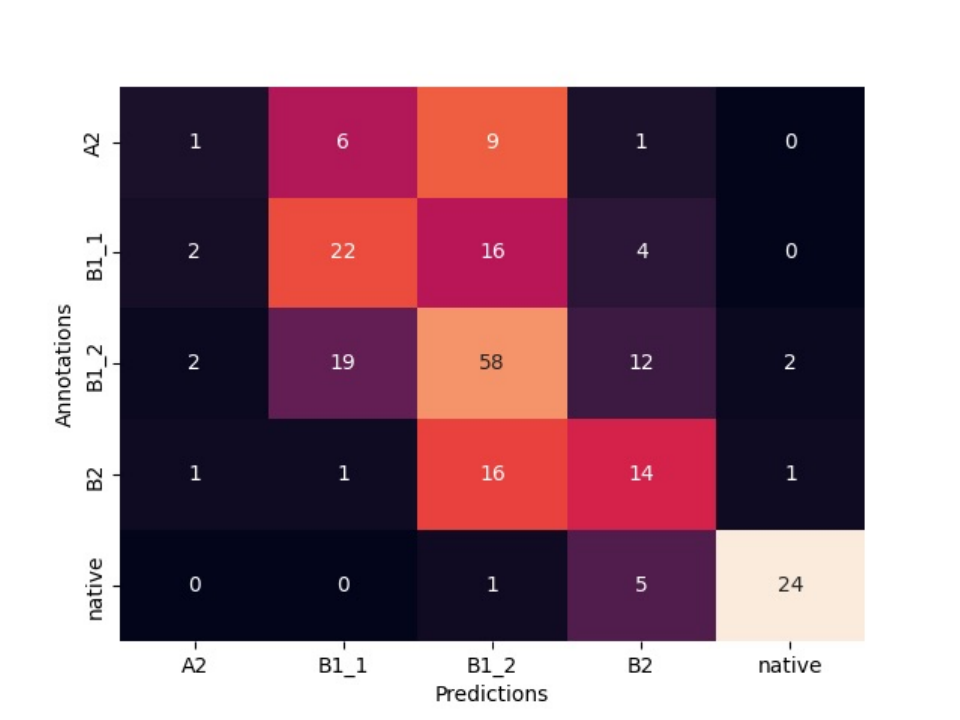}
        \caption{Focal Loss}
        \label{fig:focal-cm}
    \end{subfigure}
    \caption{Confusion Matrix on ICNALE Testing Set}
    \label{fig:confusion-matrix}
\end{figure*}

\subsection{Configurations}
We adopt a two-stage pipeline for ASA:
\begin{enumerate}
    \item Transcription Stage:
Spoken responses are first transcribed using Whisper-large-v2, a state-of-the-art ASR model known for its robustness across accents and noise conditions.
    \item Classification Stage:
The transcribed text is then fed into a BERT-base-uncased model for proficiency classification. Token-level embeddings are aggregated using mean pooling, followed by a dropout layer and a linear classification head.
\end{enumerate}

To ensure stable and effective learning, we configured training with a batch size of 4 and gradient accumulation over 2 steps (yielding an effective batch size of 8), trained for 20 epochs using a learning rate of $1 \times 10^{-5}$, and optimized with either the standard cross-entropy loss, focal loss, or the BLV loss. Where applicable, we employ mixed-precision (FP16) training and apply a 0.1 dropout rate to the final layer of the feed-forward network.

\subsection{Evaluation Metrics}

In order to validate whether our method effectively addresses the data imbalance issue in the ASA task, we employed a set of complementary evaluation metrics that capture different aspects of prediction quality.

To validate the effectiveness of our method on the ASA task, we employed Pearson Correlation Coefficient (PCC), Root Mean Square Error (RMSE), Accuracy, and Adjacent Accuracy as primary evaluation metrics. To further investigate the impact of data imbalance, we utilized macro-averaged variants of Accuracy and RMSE, which give equal weight to all classes regardless of their frequency. Additionally, the F1-score was included to balance precision and recall, particularly important for assessing performance on minority classes in imbalanced datasets. Collectively, these metrics provide a comprehensive framework to evaluate both overall prediction quality and fairness across proficiency levels.

\begin{figure*}
    \centering
    
    \begin{subfigure}{0.3\textwidth}
        \includegraphics[width=\textwidth]{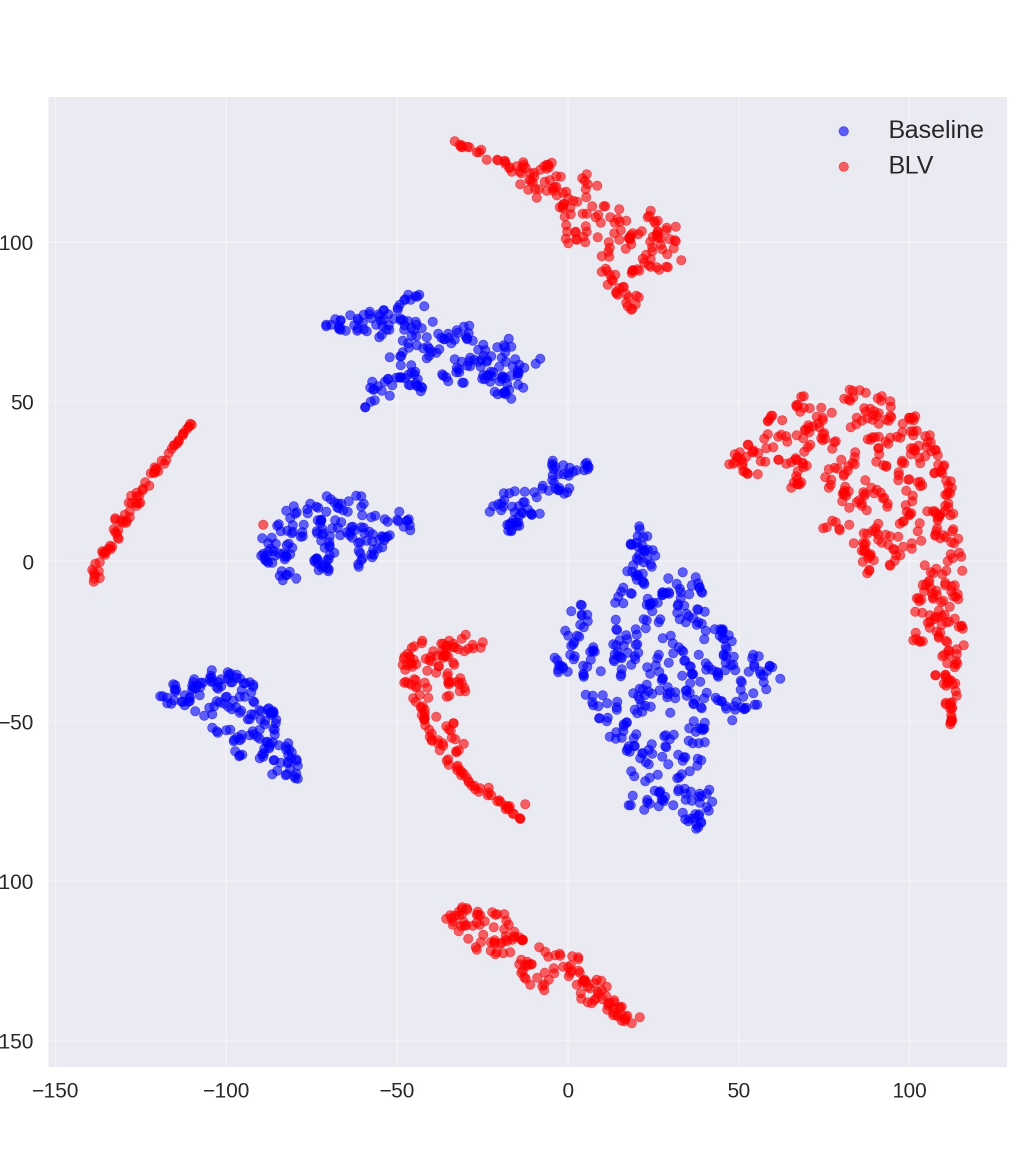}
        \caption{Baseline v.s. BLV}
        \label{fig:tsne-training-overall}
    \end{subfigure}
    \hfill
    \begin{subfigure}{0.3\textwidth}
        \includegraphics[width=\textwidth]{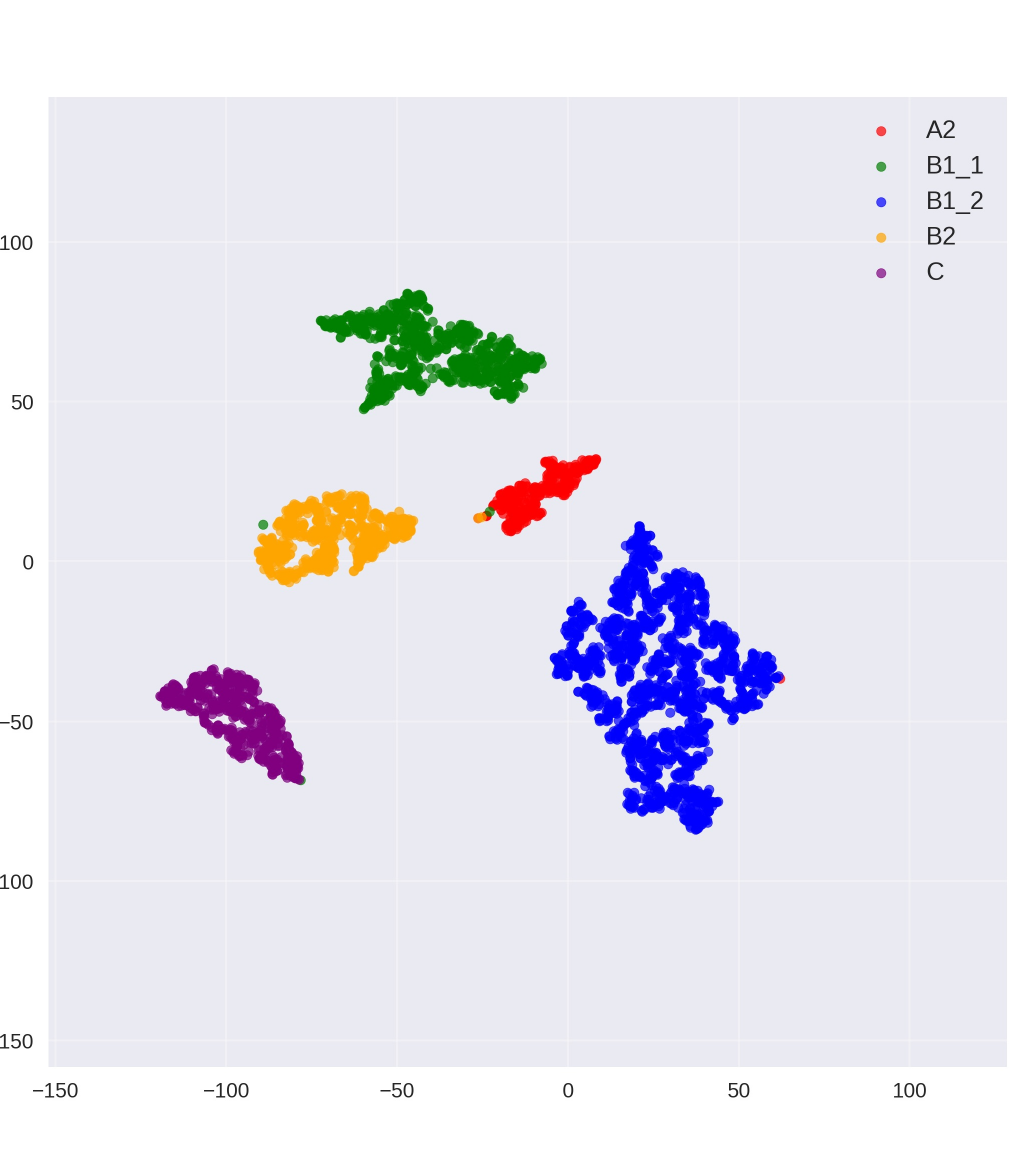}
        \caption{Baseline}
        \label{fig:tsne-training-baseline}
    \end{subfigure}
    \hfill
    \begin{subfigure}{0.3\textwidth}
        \includegraphics[width=\textwidth]{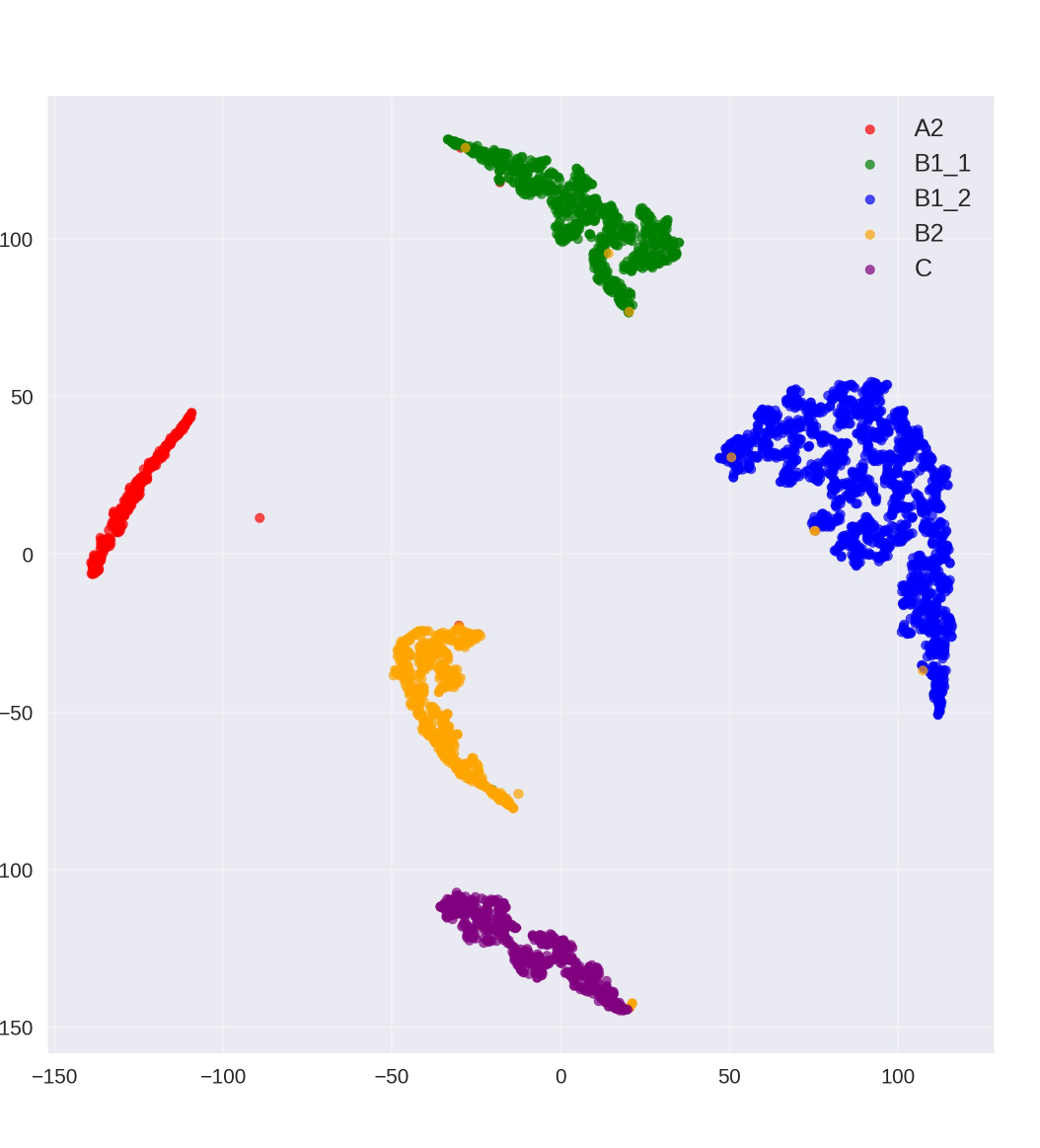}
        \caption{BLV}
        \label{fig:tsne-training-blv}
    \end{subfigure}
    \caption{t-SNE Visualization of the logits on ICNALE Training Set}
    \label{fig:tsne-training}
\end{figure*}

\section{Results}

\subsection{Main Result on ICNALE}

The results in Table \ref{tab:ICNALE result} demonstrate the effectiveness of our BLV variance injection method in improving the model’s overall performance. We observe consistent gains in PCC, RMSE, standard accuracy, and adjacent accuracy, which collectively reflect better prediction accuracy and stronger correlation between predicted and true proficiency levels.

To specifically verify the method’s impact on data imbalance, improvements in macro RMSE and macro accuracy indicate a more balanced performance across all classes, mitigating bias toward majority classes. Additionally, the increase in F1-score highlights enhanced detection of minority classes by effectively balancing precision and recall, which is critical in imbalanced classification scenarios.
Together, these metrics provide a comprehensive validation of our method’s ability to improve both overall accuracy and fairness across imbalanced proficiency levels.

Among the BLV variants tested, $\sigma=6$ outperforms $\sigma=2$, indicating that a larger variance injection plays a critical role in enhancing the model's predictive capability. This suggests that increasing the variance allows the model to better address data imbalance and learn more discriminative representations. Importantly, both $\sigma=2$ and $\sigma=6$ settings deliver clear improvements over the baseline model, highlighting the robustness and effectiveness of our BLV approach across a range of parameter values. These findings underscore the importance of carefully tuning the variance parameter to fully realize the benefits of variance injection in mitigating imbalance-related biases.

We also evaluated the BLV loss against focal loss, a widely adopted objective for addressing class imbalance. Although focal loss leads to higher F1 scores and improved macro-accuracy, this gain comes at the cost of reduced overall accuracy and PCC. The confusion matrices in Fig. \ref{fig:confusion-matrix} highlight how these two methods differ in their error distributions.

Overall, these grouped evaluation metrics provide comprehensive evidence that our method effectively mitigates the impact of data imbalance, leading to a more equitable and accurate predictive model.

\subsection{Visualization of logits on ICNALE with BLV loss Injection}

We further investigate how variance injection via BLV loss affects the model’s learned representations by comparing the output logits of the baseline BERT and our BERT + BLV $\sigma=6$ variant through t-distributed stochastic neighbor embedding (t-SNE) \cite{van2008visualizing}. We applied t-SNE with a perplexity of 30 over 5,000 iterations to project each model’s high-dimensional logit vectors into two dimensions.  As illustrated in Fig. \ref{fig:tsne-training}, the BERT + BLV embeddings exhibit substantially larger inter-cluster separations and markedly tighter intra-cluster groupings than those of the baseline.  This pronounced clustering behaviour confirms that injecting higher variance into minority-class logits enhances the model’s ability to discriminate between adjacent proficiency levels.

\section{Conclusion}

In conclusion, this study addresses the challenge of data imbalance in prediction models by introducing variance injection via BLV loss. Our method is rigorously validated across multiple evaluation metrics, providing strong evidence of its effectiveness in reducing bias and improving model robustness. Additionally, visualization of the learned representations confirms that variance injection enhances the model's discriminative ability. These findings demonstrate the potential of our approach to develop fairer and more reliable models in imbalanced data scenarios.

\printbibliography

\end{document}